\documentclass[9pt]{article}
\usepackage[margin=1.25in]{geometry}
\usepackage{graphicx} 
\usepackage[square,sort,comma,numbers]{natbib}
\usepackage{booktabs}
\usepackage{multirow}
\usepackage{enumitem}
\usepackage{authblk}

\begin{document}

\title{Tasks and Roles in Legal AI: Data Curation, Annotation, and Verification}
\author[a]{Allison Koenecke}
\author[b]{Jed Stiglitz}
\author[a]{David Mimno}
\author[a]{Matthew Wilkens}

\affil[a]{\footnotesize Department of Information Science, Cornell University, Ithaca, NY 14853}
\affil[b]{\footnotesize Law School, Cornell University, Ithaca, NY 14853}

\date{}
\maketitle

\begin{abstract}
The application of AI tools to the legal field feels natural: large legal document collections could be used with specialized AI to improve workflow efficiency for lawyers and ameliorate the “justice gap” for underserved clients. However, legal documents differ from the web-based text that underlies most AI systems. The challenges of legal AI are both specific to the legal domain, and confounded with the expectation of AI’s high performance in high-stakes settings. We identify three areas of special relevance to practitioners: data curation, data annotation, and output verification. First, it is difficult to obtain usable legal texts. Legal collections are inconsistent, analog, and scattered for reasons technical, economic, and jurisdictional. AI tools can assist document curation efforts, but the lack of existing data also limits AI performance. Second, legal data annotation typically requires significant expertise to identify complex phenomena such as modes of judicial reasoning or controlling precedents. We describe case studies of AI systems that have been developed to improve the efficiency of human annotation in legal contexts and identify areas of underperformance. Finally, AI-supported work in the law is valuable only if results are verifiable and trustworthy. We describe both the abilities of AI systems to support evaluation of their outputs, as well as new approaches to systematic evaluation of computational systems in complex domains. We call on both legal and AI practitioners to collaborate across disciplines and to release open access materials to support the development of novel, high-performing, and reliable AI tools for legal applications.
\vspace{10pt}

\emph{Keywords: Law $|$ Large Language Models $|$ Trust $|$ Algorithmic Bias $|$ Documents}
\end{abstract}

\section{Introduction}

\noindent Artificial intelligence (AI) is rapidly evolving and holds tremendous promise for transforming the legal profession and increasing access to justice. At the same time, law is a challenging domain for AI, and the consequences of errors can be significant. For many relevant individuals and contexts, the high stakes of legal judgments will be comparable only to those in healthcare. The wrong legal advice could expose a person to civil liability and severe financial consequences or, even more seriously, to criminal liability and possible loss of liberty.

There is considerable reason for excitement about the use of AI for law. Models show strong performance in information retrieval, factual question answering, and even higher-level reasoning abilities \citep{brown2020language, ouyang2022training}. Companies have particularly touted results on closed-ended evaluations like bar exams \citep{achiam2023gpt}. Recent work, however, has highlighted some of the risks of AI, particularly around trusting the test-oriented evaluations common in machine learning \cite{martinez2024re, kapoor2024promises}. As Kapoor et al. observe, ``it is not a lawyer’s job to answer bar exam questions all day.'' They argue for the use of AI in narrowly defined, low-stakes applications rather than, for example, predicting case outcomes or penalties.
The objective of this article is to lay out the practical difficulties for AI in the context of legal processes, and to suggest technically innovative and ethically responsible paths forward. 

AI has many potential applications in law, which we understand to fall into an array with \emph{who} on one dimension and \emph{what} on a second dimension. The \emph{who} dimension relates to the user of AI: one of a member of the public, a lawyer, or a judge. The \emph{what} dimension speaks to the purpose or the use of AI, and can be characterized as either a task or a judgment. Tasks might include, for example: document review, where a party is trying to find material relevant to their case in a large set of documents; legal research, where a party is trying to summarize existing law or to find the legal decisions or statutes most relevant to their case; and document drafting, where a party uses AI to create an initial draft of a letter, petition, brief, or other submission to a court or another party. Judgments, on the other hand, represent conclusions, determinations, or decisions. A judge, for instance, will decide whether the petitioner or respondent wins a case. A lawyer will conclude that their client would not or would probably not violate the law through some action. Or the client themselves might reach that judgment. AI can play a role in any of these judgments. Table \ref{tab:matrix} provides examples of how various users might employ legal AI.

\noindent\begin{table}[htbp]
\centering
\caption{Legal AI Users and Purposes}
\label{tab:matrix}
\renewcommand{\arraystretch}{1.3}
\setlist[itemize]{nosep, leftmargin=*, before=\begin{minipage}[t]{\hsize}\raggedright, after=\end{minipage}}
\begin{tabular}{p{2cm}p{4.5cm}p{5cm}}
\toprule
\textbf{User} & \textbf{Legal Tasks} & \textbf{Legal Judgments} \\
\midrule
\textbf{Judge} & 
\begin{itemize}
  \item Retrieving/summarizing legal material
  \item Drafting judicial opinions
  \item \textbf{Risk}: Over-reliance, de-training
\end{itemize} & 
\begin{itemize}
  \item Determining liability (civil/criminal)
  \item \textbf{Risk}: Judgment errors, undermining deliberative function
\end{itemize} \\
\midrule
\textbf{Lawyer} & 
\begin{itemize}
  \item Document review
  \item Retrieving/summarizing legal material
  \item Drafting contracts and briefs
  \item \textbf{Risk}: Over-reliance, de-training
\end{itemize} & 
\begin{itemize}
  \item Assessing liability
  \item Evaluating litigation prospects
  \item \textbf{Risk}: Judgment errors, failure to account for values
\end{itemize} \\
\midrule
\textbf{Public} & 
\begin{itemize}
  \item Document review
  \item Retrieving/summarizing legal material
  \item Drafting basic legal documents
  \item \textbf{Risk}: Inability to assess outputs
\end{itemize} & 
\begin{itemize}
  \item Assessing liability
  \item Evaluating litigation prospects
  \item \textbf{Risk}: Errors leading to liability, false reliance
\end{itemize} \\
\bottomrule
\end{tabular}
\end{table}

Both the challenges that AI faces, and the ethical problems presented by it, depend on what position within the \emph{who}-\emph{what} array is being considered. The technical ability of AI to render judicial decisions, as though a judge, is different from the technical ability of AI to summarize documents for a lawyer, as though it were an unbarred summer associate. And so too are the consequences of occupying either of those cells: in the first, a determination that could expose the defendant to financial burdens or loss of liberty, and in the second, a preliminary determination that might be reviewed and corrected by a domain expert (e.g., a partner in the firm). Throughout our assessment, we seek to connect the technical challenges relevant to AI to the appropriate cell of the array.

The hopeful promise of AI is that it will improve productivity in the legal sector without imperiling other values such as fairness and trust in the legal system. For civil matters, there is no constitutional right to a lawyer, leading to a gulf in the provision of legal services between higher- and lower-income citizens. A recent report by the Legal Services Corporation found that low-income Americans do not receive sufficient legal assistance in over ninety percent of their substantial civil legal problems, involving issues such as housing, healthcare, consumer protection, employment, and income maintenance \citep{legal2022justice}. It is possible to imagine AI transforming the legal market so that that justice gap narrows. However, it is also possible to imagine AI leading to more, but less responsible, provision of legal services, or to using AI to relieve judges from the pressure of their caseloads \citep{posner2025judge}, undermining trust in the legal system.	

Even as AI performs remarkably well on many sophisticated intellectual tasks and rapidly penetrates diverse areas of the economy, the legal domain is set apart~\cite{mahari2023law} despite ambitious efforts \citep{shu2024lawllm}. The first goal of this article is to explain why law is different. What about the domain makes it more difficult? And why do we think of the legal domain as different from other difficult domains, such as deriving mathematical proofs? After explaining these characteristics, we point out the foundations of the technical problems that AI encounters: data curation and availability; data annotation; and output verification. We conclude by suggesting innovative and ethical ways forward in law and AI.

We stress at the outset that our views depend on the technical affordances of current AI technology. Most contemporary AI systems rely on models initially trained on vast collections of general purpose text, which include --- but are not limited to --- law-related material, and subsequently specialized through more task-specific training data. User interactions with models are then intermediated through systems that might define additional instructions, curate inputs and outputs, or shape interaction patterns. The resulting capabilities are likely to grow over time as legally relevant data becomes more available, model architectures become more sophisticated, verification devices grow more reliable, and the most effective modes of human-AI interactions are better understood \citep{jiang2023legal}. In fact, our assessment aims to advance these capabilities responsibly.

\section*{What Makes Legal AI Different?}

Though law is different from most domains, it is similar in many respects to another high-stakes area: healthcare. Over a half century ago, Kenneth Arrow wrote a seminal article that explained why the healthcare market was different from other markets, and why healthcare markets tended to perform poorly~\cite{arrow1978uncertainty}. Our interest is in law and AI, not healthcare and markets. But there are parallels between the substantive challenges of healthcare for markets and the substantive challenges of law for AI.
 
As in healthcare, there is substantial uncertainty and complexity in law. In each area, it takes humans years of specialized study to understand the domain. Legal or healthcare problems may be hard to identify or diagnose, the most appropriate solution is often unclear and must be tailored to individual circumstances, and even the most appropriate solution does not necessarily yield certain results. That uncertainty makes it difficult to evaluate the efficacy of a service, even ex-post. Consider several of the complexities relevant to law and AI. The relationship between legal authorities is complex both temporally and institutionally. Legal texts are often inter-dependent and may be fragmented, such that the meaning of a legal provision in one legal jurisdiction depends on the meaning in another legal jurisdiction. The meaning of a statutory term, for example, may depend on the historical interpretation of it adopted by an administrative agency. Or, based on lexical similarity, a particular case may be highly salient to a legal query --- but it may be irrelevant legally because the case was overruled or limited by a more recent decision in that jurisdiction. The facts of any specific case, moreover, typically will not align perfectly with earlier cases, meaning that the most appropriate diagnosis and solution is likely to be some combination of earlier decisions, with weights based on the degree to which they are analogous.

Also as in healthcare, there are significant information asymmetries between the consumers and the providers of the service. The provider likely has a good sense of how to diagnose a problem and its most appropriate solution. But that again is only because they were screened for ability and spent years in specialized training. The consumer does not have the benefit of that training. As a result, the consumer is in a vulnerable position: they could easily be misled or taken advantage of. That low baseline of consumer information, likewise, presents problems for law and AI. It means that the consumer will likely not be able to evaluate law-related AI output. This is unlike, say, asking AI to prepare a report on the best restaurants in town, output that might be validated against the user’s experience or third-party sources. With law, the user will likely not have any first-hand experience of the legal issue and will not be trained to assess the plausibility of the output. As a result, if the user is a member of the general public, our standards for trustworthiness will be higher in law applications of AI than in other, more common domains. In both health and law, high stakes amplify concerns about these asymmetries.
 
Many of the distinctive social features of the medical and legal professions might be understood as responses to these challenges of complexity and information asymmetry. Both fields elevate non-market values and norms, such as the public welfare, fairness, justice, and prioritizing client or patient interests. Both fields invest in the trustworthiness of practitioners. Both fields invest in regulatory structures that promote those outcomes, including accreditation of specialized educational institutions and licensing of professionals. It can be debated how well these forms of social regulation address the distinctive issues of the healthcare and legal markets. But what is more clear is that AI cannot easily substitute for any of them.

\section*{Technical Problems of  Legal AI}

Legal AI encounters many similar technical problems to healthcare-related AI, but the downstream applications are sufficiently different that many lessons and best practices in the medical space cannot be ported directly to our application domain: for example, using computer vision to identify tumors on x-rays is a common medical task, but it has no direct analogue in the legal domain (which may instead involve identifying relevant words on a legal document).\footnote{Both legal and healthcare-related AI share many technical fundamentals. For example, AI models are \emph{trained} on corpora of data --- meaning that the model has learned patterns from these \emph{training data} and can use these patterns either to make predictions (\emph{inferences}) about new data, or to generate new data (e.g., ChatGPT is a \emph{generative model}). Practitioners often use large \emph{pre-trained models}, which are out-of-the-box models that have already been trained on large amounts of data and that can be applied directly to new data. However, domain experts in niche fields will often \emph{fine-tune} existing pre-trained models to improve model performance on their specific task (which might involve specific types of data or ideal outputs).}

Instead, we propose three primary challenges faced by text-based legal AI technology: data curation, data annotation, and output verification. These three challenges are inherently intertwined as part of an AI pipeline: the curation of data directly informs what data are used as the basis for AI-based annotations, these annotations may be used to verify AI-generated outputs, and the AI-generated outputs may further be used to update the data curation process. Challenges faced at any point in the pipeline could render the overall AI product unusable. We discuss each challenge in the three sections below.

\section*{Challenge 1: Data Curation}

The most immediate challenge facing researchers dealing with legal documents is the ability to access sufficient quantities of documents in high-quality, machine-readable formats. Simply achieving any access to legal documents can be a hurdle. Though legal documents are typically public records, acquiring copies of documents is not always straightforward. Documents are often collected by aggregators and for-profit databases (e.g., PACER), which do not have incentives to make large-scale downloading simple. Relevant documents can also be fragmented between multiple data providers and jurisdictions. Knowing how to find documents, and which services carry which sources, is itself a skill. Obtaining usable legal texts is therefore difficult (and costly) for AI researchers and legal scholars. While quantitative AI methods can be applied to smaller datasets,\footnote{AI tools can assist document curation efforts, but the lack of existing data also limits AI performance.} the advantage of AI approaches is more clear in cases where the volume of material is beyond the scope that can be easily read and analyzed by humans.

Beyond data quantity, another consistent issue is data quality. Algorithms require clean text in accessible formats. While a lawyer can read a blurry scan of a poorly printed legal document, a computer generally cannot. Legal documents are often created and distributed by courts and other government agencies that do not have robust IT budgets or strong technical expertise. While some jurisdictions produce high-quality, born-digital documents, others provide raster-image scans of inconsistently printed documents. Extracting clean, usable text through optical character recognition (OCR) and other machine vision tools can be challenging for several reasons~\cite{piper2020page}. These include standard OCR problems like skewed page images, but also formatting problems specific to legal text, such as line numbers that can be mixed into text, and speaker attribution and disambiguation for trial transcripts. For example, a complete utterance by a prosecutor may be unintentionally split into multiple sections, some or all of which may be incorrectly attributed to a different speaker. Furthermore, the collection of underlying data in AI pipelines may exacerbate existing human biases: for example, if an AI decision tool is trained on case rulings wherein judges historically exhibited biases against certain types of defendants (whether across the board or because documents from such jurisdictions were more easily obtained), then future judges using this AI decision tool could implicitly make similarly biased judgments, inducing a feedback loop of biases stemming from biased data curation.

Handling legal documents may require more care than other types of text data. While there has been considerable interest and scholarship on copyright and intellectual property in the area of AI and law \citep{palace2019if, hugenholtz2021copyright}, lawyers have additional professional responsibilities to serve their clients, including duties of competent representation, candor, and confidentiality \citep{walters2018model, hatfield2019professionally, wendel2019promise, yamane2020artificial}. The standards of legal ethics may ask for greater restriction on the use and distribution of certain documents. This limitation is of particular concern for AI systems, which often use cloud computing and commercial APIs. While the extent to which companies like Google, Anthropic, or OpenAI are actively storing user data and training models on it is uncertain, there is no way to know or control what they are doing with uploaded documents. Moreover, the need for access to high-quality text and user interaction data to train powerful models in a competitive AI business-services environment provides a clear business and technical rationale for using such data, one that is only counteracted by companies' reputational concerns and technical data processing abilities. Concerns about the leakage of private law-related user data may be of particular relevance to the broader public.

In terms of copyright, legal documents are more likely to be in the public domain than are other varieties of text, but there are complexities. While documents themselves may be public domain, \emph{annotations} on documents and commentary about documents can be individual creative works. These additional layers of text may add significant value to documents, but also complicate questions of ownership.\footnote{An early line of cases considered whether the official reporters could claim copyright to judicial opinions. The Supreme Court clarified that they could not claim a copyright interest in judicial opinions, but that they did have an interest in accompanying explanatory materials, annotations, and headnotes they produced. Callaghan v. Myers, 128 U.S. 617 (1888). A more recent Supreme Court decision further clarified that annotations produced by a non-public actor but commissioned by the state according to its specifications would also fall outside copyright interest. Georgia v. Public.Resource.Org, Inc., 140 S.Ct. 1598 (2020). 
} A key legal question is how ``transformational'' the AI is of the copyrighted material.\footnote{For a recent case on this topic, see Thomson Reuters v. Ross Intelligence Inc., 694 F. Supp. 3d 467 (D. Del. 2023) (denying motion for summary judgment, holding that fair use defense and other matters should go before a jury).
} Such concerns may be most relevant for lawyers: models are often unable to fully “forget” copyright materials previously included in the training data, despite post-hoc methods to prevent generating protected content~\cite{wei2024evaluating}.

Finally, besides the quality of text documents themselves, metadata is of vital importance. Information such as the date of a ruling, the jurisdiction, and relations to other rulings that may supersede a given document are critical. This information is required for high-level legal reasoning: a case that is more relevant to a query on the basis of its text may be less valuable if it has been limited or overturned by later rulings. While managing metadata is not intrinsically difficult, it is complex and requires persistence, coordination among stakeholders, and detail-oriented annotation.

Addressing these issues will require a combination of technical and organizational changes. 
Better OCR correction may make lower-quality documents more accessible, or at least less noisy.
But much of the difficulty of accessing large-scale document collections is due to stubbornly human problems like uncoordinated, distributed document archiving and proprietary databases.
Perhaps the most significant changes may be due to a growing sense of the value of collecting large volumes of text for the purposes of AI training, as opposed to traditional ``one document at a time'' reading patterns.
It may happen, however, that even if large-scale, high-quality document aggregation exists, it may be restricted only to those able to pay premium prices.

\section*{Challenge 2: Data Annotation}

A commonly desired use for AI systems in legal contexts is document \emph{annotation} and \emph{classification}. Annotations range from the simple (identifying word-level named entities such as people and organizations, or tagging citations; these are likely to fall under the ``legal tasks'' heading in Table \ref{tab:matrix}) to the complex (differentiating modes of judicial reasoning, identifying contract provisions that require specialized review, or determining controlling precedents; ``legal judgments''). Classifications are functionally equivalent to annotations, though the term is often used to describe labels applied to entire documents or to significant subsections thereof. Although annotation processes may not resemble the interactive AI conversations with which end users are familiar, they are central to lawyers’ work and straightforwardly amenable to AI applications.

A subset of simple annotations, especially in cases where legal text closely resembles ordinary-language web text, are characterized by high performance and relatively low error risk. Such tasks may include, for example, identifying organizations or subject matter in discovery documents, or sorting cases by their broad legal area \cite{darji2024challengesconsiderationsannotatinglegal}. But more complex labeling tasks are generally both the most risky (because they are consequential and difficult to detect) and are those with which general-purpose AI systems struggle. 

Complex annotation tasks challenge AI systems for two primary reasons. On one hand, they are often difficult even for highly trained human readers. They may involve concepts that are ambiguous or contested by experts, or that are suffused throughout a document without being strictly localizable to a single word, sentence, or paragraph, or that depend on the correct application of detailed knowledge not explicitly indicated in the object text. At the same time, complex, domain-specific annotations are unlikely to be well represented among the training data for AI models, further lowering the performance of those models on such labeling tasks even beyond that predicted by their inherent difficulty, since system performance generally decreases on out-of-distribution data.

If pretrained generative models perform poorly on complex in-context legal learning tasks, an alternative is to finetune those models on novel expert-derived annotations. That is, pretrained models can be adapted to reproduce expert judgments on a specific task learned from a set of examples that, while not small, is a tiny fraction of the size of the model’s original training data. Still, these examples are costly and time-consuming to create, since they can be produced only by experts. Moreover, the difficulty in many cases of achieving consensus among expert annotators means that multiple annotators must label each training instance, and their disagreements must be reconciled. For these reasons, high-performing legal annotation models are typically expensive to produce, limiting their potential to narrow resource-based justice gaps.

Recent research on legal annotation illustrates these difficulties and provides several sources of optimism. Wen-Yi et al.\ find that AI annotation systems are primarily useful as aids to human annotators in high-stakes legal settings such as death-penalty trials, rather than as components of a fully automated workflow~\cite{wen2024automate}. Thalken et al.\ and Stiglitz and Thalken evaluate the ability of AI systems to identify modes of legal reasoning in opinions of the U.S. Supreme Court, finding that then-state-of-the-art (SOTA) generative models such as GPT-4 and Llama-2 struggle with the task, even when provided with instructions equivalent to those developed for human labelers~\cite{thalken-etal-2023-modeling, Stiglitz_2024}. In contrast to many results on less complex reasoning tasks, Thalken et al.\ achieve near-human-level performance using a smaller model (LEGAL-BERT) fine-tuned on the task and pretrained on in-domain data. This finding is sobering, because it suggests that complex legal-domain tasks may not be straightforwardly addressed by those AI systems (such as general-purpose chatbots) characterized by the lowest technical barriers to use. But the relatively high performance of the aging LEGAL-BERT architecture also provides evidence that, in cases where a reasonable set of export-sourced training data can be produced, even very difficult legal annotation tasks can be tractable with limited computational demands.

There are several methods by which the quality and cost of AI-generated legal annotations might be improved. As noted, finetuning on expert annotations can produce large improvements in labeling performance in cases where suitable training data can be obtained. This is true even when applied to smaller models, which are typically easier and less expensive to train, tune, and apply. In the absence of such tuning data, labeling tasks can sometimes be decomposed in a series of steps, some of which steps achieve acceptable performance even in the absence of substantial finetuning. In such cases, AI systems may be used to complete the easier steps, reclaiming expert time to focus on more difficult steps and thus lowering the total cost of the annotation pipeline. SOTA reasoning models may perform better on the reasoning tasks that legal annotations of complex concepts typically require, though the extent of any such improvement has not yet been adequately evaluated.

Finally, we note that, especially but not exclusively in academic contexts, a potential benefit of the systematic investigation of AI-assisted labeling performance is greater clarity and specificity concerning the underlying legal concepts that such labeling requires of its practitioners. In cases where an adequate specification of the criteria by which a label or annotation is to be applied cannot be agreed among experts, and, in consequence, consistent human annotation across multiple coders is impossible, the concept itself may benefit from further investigation. While the need for adequate conceptual specificity in the law is not novel, the challenges associated with implementing AI-based labeling systems may have the unintended benefit of identifying problematically unclear statutes, concepts, and definitions.

\section*{Challenge 3: Output Verification}

The third challenge we identify concerns the difficulty in verifying the quality of AI-based outputs. This challenge is, of course, not limited to AI tools: human experts are fallible, and qualitative work (especially in adversarial environments like law) is broadly challenging to benchmark with quantitative metrics. The metrics for verifying outputs may also differ by task; relevant characteristics against which to measure outputs can involve accuracy, consistency, groundedness, and so on.\footnote{There are many related outputs that differ slightly in measurement and are also commonly used, such as: reliability, trustworthiness, alignment, and reproducibility.}

\paragraph{Accuracy} An example of an AI-based legal task primarily measured for accuracy is the use of speech-to-text systems for generating transcriptions, whether in the courtroom, in meetings for notetaking purposes, or from bodycam footage for generating police reports (e.g., Axon's FirstDraft~\cite{murphy2024police}). This task is of importance for legal teams and defendants alike. Such systems can be measured for accuracy by comparing machine transcriptions to the \emph{ground truth} of what was truly said (usually generated by human experts, such as court reporters). One common metric, the \emph{Word Error Rate} (WER), counts the number of word insertions, deletions, or substitutions between the ground truth and machine transcriptions, and normalizes by the number of words in the ground truth. In audits of AI-based speech-to-text systems, WER is straightforward to calculate programmatically, and reporting this metric can inform practitioners of which commercial providers of the AI service perform best at transcription~\cite{zhao2025quantification}. However, the reported WER from external audits may not be reflective of the most relevant benchmarks to the legal task at hand (e.g., audits of services may be for general speech, and may mask underperformance on transcriptions of niche legal terms). Furthermore, WER may differ for different user demographics: lower accuracy has been found for speakers of African American English~\cite{koenecke2020racial} and for speakers with language or hearing disorders~\cite{koenecke2024careless}. As such, the onus is on the legal practitioner to decide what levels of accuracy may be acceptable for the AI tool, especially when considering the relevant client demographics.

\paragraph{Consistency} A common task for lawyers requiring output consistency regards labeling massive document collections, as described in the previous section. In the past, participants had to rely on brittle quantitative tools (e.g., inter-annotator agreement metrics~\cite{wacholder2014annotating}) to confirm that multiple human annotators agree with each others’ labels. However, AI tools allow us to scale up the task of labeling: not replacing humans, but allowing for additional signal about consistency and confidence of labels. This could still be measured through inter-annotator agreement (now at scale, between humans and a ``jury'' of machine-predicted labels~\cite{gordon2022jury}), but can also be visualized through machine outputs containing error bars or other indications of the degree of potential variance in labels. The exercise of uncovering inconsistent labels is not only a quantitative one. Discrepancies between AI outputs and human annotators also provide annotators with the opportunity to reflect qualitatively on the object of their annotations, and to provide feedback for updating the AI tool as necessary~\cite{wen2024automate}. Consistency may also be of interest to judges using AI tools to ensure that they are not perpetuating biases across cases. For example, in orders concerning pre-trial detention, judges have benefited from Praxis, a decision grid that maps algorithmic risk score (VPRAI) and charge type to recommended decision (level of supervision). Despite the fact that the algorithm itself was imperfect, Praxis training affected decision-maker behavior by increasing consistency with the Praxis recommendation: pre-trial officers in a randomized controlled trial were 2.3 times more likely to recommend release after undergoing Praxis training compared to those who did not receive Praxis training; judges released Praxis group defendants at first appearance 1.9 times more often compared to defendants not assigned to Praxis; and, defendants in Praxis groups were 1.3 times less likely to experience a new re-arrest or failure to appear~\cite{danner2015risk}.

\paragraph{Groundedness} Legal research tools may increasingly rely on large language models (LLMs, used interchangeably with AI) to perform tasks like document summarization and document drafting. While such AI tools may not produce output that seems significantly different from an inexperienced research associate, they run the risk of appearing more confident in their outputs, and generating ungrounded statements in ways that are harder to detect. In particular, there has been significant documented risk of LLM hallucinations, defined as generated text ``that is nonsensical, or unfaithful to the provided source input''~\cite{ji2023survey}. Tools produced by LexisNexis and Thomson Reuters were found to hallucinate 17--33\% of the time~\cite{magesh2024hallucination}. This production of ungrounded text has massive downstream repercussions, such as sanctions of lawyers who have not appropriately fact-checked AI-generated text for hallucinated cases, quotes, or citations~\cite{weiser2023here}. As of this writing, the technology does not yet exist to predict definitively whether a text includes hallucinations. As such, practitioners themselves must rely on other methods to avoid such concerns. Mitigation strategies may range from using Retrieval-Augmented Generation (RAG) models that only fetch direct quotes from a set of sources (rather than generating entirely new text), or ensuring that a pipeline exists for human expert--based verification of generated facts. The general public similarly faces groundedness concerns when using models such as ChatGPT or specialized services such as DoNotPay (formerly branded as ``the world’s first robot lawyer'') for legal advice. Hallucinations from these models may not appear obviously false to untrained users; the FTC even took action against DoNotPay for failing to serve as a substitute to a human expert~\cite{ftc2024donotpay}. 

In general, AI-supported work in the law is valuable only if results are trustworthy. Trust encompasses metrics such as accuracy, consistency, and groundedness. While the previous paragraphs describe contexts in which humans verify AI-based outputs, we note that it is also possible to use AI tools to verify human-generated outputs. For example, algorithms have been used to quantify disparities in policing decisions of traffic stops~\cite{pierson2020large}, to better understand political stereotyping~\cite{zamfirescu2022trucks}, and to capture polarization in congressional speeches~\cite{card2022computational}. Over time, advances in AI-based algorithms will unlock new approaches to systematic evaluations of complex systems in the legal domain.

We believe that output verification will require both (a) additional auditing of AI systems, and (b) keeping humans in the loop. 

Conducting audits of AI systems --- ideally across a suite of metrics~\cite{wang2024benchmark} that may address accuracy, consistency, and groundedness as is relevant to the task at hand --- allows practitioners to better understand the limitations of the AI system for that task. An open question may remain: when is a system good enough to use? The baseline may be different for different tasks. In the case of court reporting, prior work has found that, just as AI-based speech-to-text systems perform worse at transcribing African American English, human court reporters similarly underperform for Black speakers~\cite{jones2019testifying}. In the case of identifying hallucinated content generated by AI, humans may struggle as much as do advanced technical methods~\cite{magesh2024hallucination}. We may expect AI (trained on massive corpora spanning many resources) to outperform human experts, but this will not always be the case (especially for groundedness concerns such as hallucinations, or for cases where the AI model is trained on relatively little data, such as with niche annotation tasks). As such, practitioners must decide for themselves whether the limitations revealed from an AI audit support using using the tool at all.

In many cases, it may be difficult to decide whether to use AI tools compared to a human baseline. AI tools may perform better on average but be especially bad at edge cases, for example. These are precisely the set of cases where we recommend that human experts fill these important knowledge gaps. While AI tools can lessen the administrative burden placed on users for mundane tasks that can be automated with high accuracy, humans should be kept in the loop~\cite{alumae2025striving} to act as guardrails to avoid real-time AI-generated errors. In the example of speech-to-text transcriptions, human experts could listen to recorded audio to confirm machine-generated transcriptions are accurate. In legal writing, human experts can be deployed to fact-check quotes and citations to ensure they are grounded in the relevant literature. This expert-based fact-checking can serve as net positive for all parties: end users (such as the public) are shown output confirmed to be correct, mitigating the chance of harm from AI-induced errors in court. Lawyers have verified for themselves that AI-produced content is correct, mitigating the risk of  professional sanctions. And judges can make legal judgments without fear of basing a case on untruths that, if left unchecked, could have serious ramifications in future related cases. In many instances, the additional cost of validating AI outputs pales in comparison to the potential fees and fines arising from AI-based errors.

\section*{The Path Forward for Legal AI}

Law, like healthcare, is a highly regulated environment that is substantially more complicated than many current AI application settings.
Conclusions are contextual and cannot be easily drawn from single documents, distinguishing \emph{true} from merely \emph{plausible} requires years of training and experience, and the consequences of errors and misinterpretations can be literally life or death.
The law cannot be ``fixed'' by the mere application of AI.
Our best approach is to identify specific challenges that are both relevant for actual legal practice and amenable to careful and limited design and evaluation.

\paragraph{Ranking and comparison of challenges} 
Of the three challenges we have identified --- data curation, document annotation, and output validation --- we judge output validation to be both the most difficult and the the most important.

To be clear, all three areas present central challenges for the development and appropriate use of AI in legal settings. But the lack of suitable data or adequate labeling performance present comparatively clear failure modes. It would be unfortunate and, perhaps, a matter of bias if documents related to a given jurisdiction or area of law are unavailable for use with AI systems. But it will generally be clear to users that they have encountered this limitation, allowing them to act appropriately. Likewise, poor labeling performance will stand in the way of important legal work. AI will not be as useful as it might be. But if it is known that the annotation is incorrect, the user can discount the AI response. 

In isolation, the problems of data curation and poor labeling performance hinder the usefulness of legal AI. But they do not render legal AI untrustworthy --- merely limited.

Output verification is different. Without a sense of whether the output is verified or reliable, legal AI remains untrustworthy. In the world of software, verification involves a search for what are sometimes called \emph{logic errors}. When a program runs and produces incorrect but facially reasonable outputs, it can be difficult even to recognize than an error has occurred. Logic bugs have persisted in software and hardware systems despite extensive testing, only to cause catastrophic failures in later use.\footnote{An example of this kind of latent bug might be the Day crash bug, which crashed Windows 95 and 98 computers after 49.7 days ($2^{32}$ milliseconds) of continuous use due to a memory overflow issue~\cite{CNET1999}.} The risk for AI is analogous, though it is likely to exist not only in edge cases but also in the typical case. Because AI generally produces fluent natural-language output, and because ground truth for the most complex and important legal tasks is difficult to establish, the risk of over-reliance on flawed AI systems is high. This point is exacerbated by economically rational use cases that emphasize the time-saving benefits of AI systems; if a busy lawyer turns to AI to reduce their workload, they may not add back the time necessary to evaluate carefully the system they use. In the current state of AI development, there is an ever-present risk of unseen, catastrophic failure. 

\paragraph{Positioning of AI in legal workflows.} 
The primacy of output validation shapes our analysis of how AI ought to be positioned in legal workflows. A conventional wisdom --- one we have identified and endorsed in this article --- is that it is important to keep a human in the loop. Yet humans vary in their ability to supervise and evaluate model performance, and  models vary in their ability to perform different legal tasks. That insight motivates our use-task matrix for legal AI (Table \ref{tab:matrix}) --- each element in the matrix reflects a different set of risks and suggests a positioning of AI in the legal workflow.

The risks of legal AI loom large when the public is the user. The public lacks legal training and cannot reliably evaluate model output. This is a significant risk even on more routine legal tasks, such as retrieving or summarizing legal material --- the kind of relevance and summarization tasks on which language models typically perform well. These concerns are even more pronounced when it comes to sophisticated legal roles that call on models to make legal judgments, such as whether a certain action would expose a person to civil or criminal liability. For these reasons, our view is that legal AI should not generally be public- or client-facing absent robust verification solutions.

At the other end of the user spectrum, we also see concerns with judges’ use of AI for legal tasks. On routine legal tasks, though judges would be able to evaluate the relevance of retrieved material or the quality of a summary, retrieving or summarizing the material may be part of the judicial function itself. That is, sifting through legal material, identifying what is and is not relevant, coming to a view about what makes it relevant or not, and synthesizing the material may be part of the deliberative process. Without that spadework, judges may make hastier, less considered judgments that over time erode public trust in the judicial system. That conclusion is even stronger when it comes to relying on AI to render judgments, e.g., determining whether a party is civilly liable. Though the judge is legally capable, if they use AI, they will not have done the work necessary to know whether, fully considered, they believe the result to be correct. Moreover, unlike simpler legal tasks, legal judgments call on the judge to make hard tradeoffs and evaluate conduct considering human norms and values. It is not clear AI can operate on those registers \cite{posner2025judge}. Judicial use of AI, too, we believe should be viewed skeptically.

The greatest potential for legal AI may be with the user class between the public and the judge --- that is, the lawyer. Like the judge, the trained lawyer will be able to evaluate the output from the models. If the model retrieves legal documents from other jurisdictions, for example, the lawyer will be able to identify that issue. Moreover, though there is a risk of under-deliberation analogous to that of the judge, it is attenuated by forms of accountability largely silent for judges: the need to perform for clients and to maintain standing before judges themselves. Also unlike the judge, practicing lawyers often already embed in large team-production units, in which lawyers perform different tasks based on hierarchy and career stage. The human analogy for AI in this context might be an unusually quick, efficient, and inexpensive first-year associate who tends to be somewhat error-prone. That is a labor unit that law firms can likely responsibly metabolize, plausibly improving lawyer productivity and, possibly, access to justice. In line with this view, a recent randomized control trial found that the latest generation of models significantly increased lawyer productivity and work quality \citep{schwarcz2025ai}.

\paragraph{Calls to Action} We call on practitioners from both AI and legal domains to work across disciplines to mitigate potential legal AI harms. Towards challenges in data curation, legal experts can collect diverse text data that would otherwise be difficult and costly to obtain through public repositories, while AI developers can work to provide software fine-tuned on these data to address specific needs of legal practitioners. To address challenges in data annotation for legal tasks, we encourage model designers to work directly with teams manually generating annotations to better understand the nuanced considerations and workflows used by experts to perform tasks that, instead of being fully outsourced to AI, may be best performed with AI-in-the-loop as part of existing pipelines. Finally, to ensure the feasibility of output verification, the free public release of both text corpora (from legal practitioners) and AI models and code (from AI practitioners) will allow for more transparent development of legal AI, and foster the ability to perform audits of models to ensure adequate performance before being used in high-stakes scenarios.


\bibliographystyle{acm}
\bibliography{pub}

\begin{thebibliography}{10}

\bibitem{achiam2023gpt}
{\sc Achiam, J., Adler, S., Agarwal, S., Ahmad, L., Akkaya, I., Aleman, F.~L., Almeida, D., Altenschmidt, J., Altman, S., Anadkat, S., et~al.}
\newblock Gpt-4 technical report.
\newblock Tech. rep., arXiv preprint arXiv:2303.08774, 2023.

\bibitem{alumae2025striving}
{\sc Alum{\"a}e, T., and Koenecke, A.}
\newblock Striving for open-source and equitable speech-to-speech translation.
\newblock {\em Nature 637\/} (2025).

\bibitem{arrow1978uncertainty}
{\sc Arrow, K.~J.}
\newblock Uncertainty and the welfare economics of medical care.
\newblock In {\em Uncertainty in economics}. Elsevier, 1978, pp.~345--375.

\bibitem{brown2020language}
{\sc Brown, T., Mann, B., Ryder, N., Subbiah, M., Kaplan, J.~D., Dhariwal, P., Neelakantan, A., Shyam, P., Sastry, G., Askell, A., et~al.}
\newblock Language models are few-shot learners.
\newblock {\em Advances in neural information processing systems 33\/} (2020), 1877--1901.

\bibitem{card2022computational}
{\sc Card, D., Chang, S., Becker, C., Mendelsohn, J., Voigt, R., Boustan, L., Abramitzky, R., and Jurafsky, D.}
\newblock Computational analysis of 140 years of us political speeches reveals more positive but increasingly polarized framing of immigration.
\newblock {\em Proceedings of the National Academy of Sciences 119}, 31 (2022), e2120510119.

\bibitem{CNET1999}
{\sc CNET}.
\newblock Windows may crash after 49.7 days.
\newblock {\em CNET Culture\/} (1999).

\bibitem{legal2022justice}
{\sc Corporation, L.~S.}
\newblock {\em The Justice Gap: The Unmet Civil Legal Needs of Low Income Americans}.
\newblock Slosar Research LLC, 2022.

\bibitem{danner2015risk}
{\sc Danner, M.~J., VanNostrand, M., and Spruce, L.~M.}
\newblock Risk-based pretrial release recommendation and supervision guidelines.
\newblock Tech. rep., Virginia Department of Criminal Justice Services, 2015.

\bibitem{darji2024challengesconsiderationsannotatinglegal}
{\sc Darji, H., Mitrović, J., and Granitzer, M.}
\newblock Challenges and considerations in annotating legal data: A comprehensive overview, 2024.

\bibitem{ftc2024donotpay}
{\sc FTC}.
\newblock Donotpay: Case summary.
\newblock Tech. rep., Federal Trade Commission, Cases and Proceedings, 2024.

\bibitem{gordon2022jury}
{\sc Gordon, M.~L., Lam, M.~S., Park, J.~S., Patel, K., Hancock, J., Hashimoto, T., and Bernstein, M.~S.}
\newblock Jury learning: Integrating dissenting voices into machine learning models.
\newblock In {\em Proceedings of the 2022 CHI Conference on Human Factors in Computing Systems\/} (2022), pp.~1--19.

\bibitem{hatfield2019professionally}
{\sc Hatfield, M.}
\newblock Professionally responsible artificial intelligence.
\newblock {\em Ariz. St. LJ 51\/} (2019), 1057.

\bibitem{hugenholtz2021copyright}
{\sc Hugenholtz, P.~B., and Quintais, J.~P.}
\newblock Copyright and artificial creation: Does eu copyright law protect ai-assisted output?
\newblock {\em IIC-International Review of Intellectual Property and Competition Law 52}, 9 (2021), 1190--1216.

\bibitem{ji2023survey}
{\sc Ji, Z., Lee, N., Frieske, R., Yu, T., Su, D., Xu, Y., Ishii, E., Bang, Y.~J., Madotto, A., and Fung, P.}
\newblock Survey of hallucination in natural language generation.
\newblock {\em ACM Computing Surveys 55}, 12 (2023), 1--38.

\bibitem{jiang2023legal}
{\sc Jiang, C., and Yang, X.}
\newblock Legal syllogism prompting: Teaching large language models for legal judgment prediction.
\newblock In {\em Proceedings of the nineteenth international conference on artificial intelligence and law\/} (2023), pp.~417--421.

\bibitem{jones2019testifying}
{\sc Jones, T., Kalbfeld, J.~R., Hancock, R., and Clark, R.}
\newblock Testifying while black: An experimental study of court reporter accuracy in transcription of african american english.
\newblock {\em Language 95}, 2 (2019), e216--e252.

\bibitem{kapoor2024promises}
{\sc Kapoor, S., Henderson, P., and Narayanan, A.}
\newblock Promises and pitfalls of artificial intelligence for legal applications.
\newblock {\em Journal of Cross-disciplinary Research in Computational Law 2}, 2 (2024).

\bibitem{koenecke2024careless}
{\sc Koenecke, A., Choi, A. S.~G., Mei, K.~X., Schellmann, H., and Sloane, M.}
\newblock Careless whisper: Speech-to-text hallucination harms.
\newblock In {\em The 2024 ACM Conference on Fairness, Accountability, and Transparency\/} (2024), pp.~1672--1681.

\bibitem{koenecke2020racial}
{\sc Koenecke, A., Nam, A., Lake, E., Nudell, J., Quartey, M., Mengesha, Z., Toups, C., Rickford, J.~R., Jurafsky, D., and Goel, S.}
\newblock Racial disparities in automated speech recognition.
\newblock {\em Proceedings of the national academy of sciences 117}, 14 (2020), 7684--7689.

\bibitem{magesh2024hallucination}
{\sc Magesh, V., Surani, F., Dahl, M., Suzgun, M., Manning, C.~D., and Ho, D.~E.}
\newblock Hallucination-free? assessing the reliability of leading ai legal research tools.
\newblock Tech. rep., arXiv preprint arXiv:2405.20362, 2024.

\bibitem{mahari2023law}
{\sc Mahari, R., Stammbach, D., Ash, E., and Pentland, A.}
\newblock The law and {NLP}: Bridging disciplinary disconnects.
\newblock In {\em Findings of the Association for Computational Linguistics: EMNLP 2023\/} (Singapore, Dec. 2023), H.~Bouamor, J.~Pino, and K.~Bali, Eds., Association for Computational Linguistics, pp.~3445--3454.

\bibitem{martinez2024re}
{\sc Mart{\'\i}nez, E.}
\newblock Re-evaluating gpt-4’s bar exam performance.
\newblock {\em Artificial intelligence and law\/} (2024), 1--24.

\bibitem{murphy2024police}
{\sc Murphy, S., and O'Brien, M.}
\newblock Police officers are starting to use ai chatbots to write crime reports. {W}ill they hold up in court?, 2024.

\bibitem{ouyang2022training}
{\sc Ouyang, L., Wu, J., Jiang, X., Almeida, D., Wainwright, C., Mishkin, P., Zhang, C., Agarwal, S., Slama, K., Ray, A., et~al.}
\newblock Training language models to follow instructions with human feedback.
\newblock {\em Advances in neural information processing systems 35\/} (2022), 27730--27744.

\bibitem{palace2019if}
{\sc Palace, V.~M.}
\newblock What if artificial intelligence wrote this: Artificial intelligence and copyright law.
\newblock {\em Fla. L. Rev. 71\/} (2019), 217.

\bibitem{pierson2020large}
{\sc Pierson, E., Simoiu, C., Overgoor, J., Corbett-Davies, S., Jenson, D., Shoemaker, A., Ramachandran, V., Barghouty, P., Phillips, C., Shroff, R., et~al.}
\newblock A large-scale analysis of racial disparities in police stops across the united states.
\newblock {\em Nature human behaviour 4}, 7 (2020), 736--745.

\bibitem{piper2020page}
{\sc Piper, A., Wellmon, C., and Cheriet, M.}
\newblock The page image: Towards a visual history of digital documents.
\newblock {\em Book History 23}, 1 (2020), 365--397.

\bibitem{posner2025judge}
{\sc Posner, E.~A., and Saran, S.}
\newblock Judge ai: Assessing large language models in judicial decision-making.
\newblock Tech. Rep. 2503, University of Chicago Coase-Sandor Institute for Law \& Economics Research Paper, 2025.

\bibitem{schwarcz2025ai}
{\sc Schwarcz, D., Manning, S., Barry, P., Cleveland, D.~R., Prescott, J., and Rich, B.}
\newblock Ai-powered lawyering: {AI} reasoning models, retrieval augmented generation, and the future of legal practice.
\newblock Tech. Rep. 25-16, Minnesota Legal Studies Research Paper, 2025.

\bibitem{shu2024lawllm}
{\sc Shu, D., Zhao, H., Liu, X., Demeter, D., Du, M., and Zhang, Y.}
\newblock Lawllm: Law large language model for the us legal system.
\newblock In {\em Proceedings of the 33rd ACM International Conference on Information and Knowledge Management\/} (2024), pp.~4882--4889.

\bibitem{Stiglitz_2024}
{\sc Stiglitz, E., and Thalken, E.}
\newblock Historical trends in macro-jurisprudence: A language model assessment, 1870-2023.
\newblock {\em Maryland Law Review 84}, 1 (2024).

\bibitem{thalken-etal-2023-modeling}
{\sc Thalken, R., Stiglitz, E., Mimno, D., and Wilkens, M.}
\newblock Modeling legal reasoning: {LM} annotation at the edge of human agreement.
\newblock In {\em Proceedings of the 2023 Conference on Empirical Methods in Natural Language Processing\/} (Singapore, Dec. 2023), H.~Bouamor, J.~Pino, and K.~Bali, Eds., Association for Computational Linguistics, pp.~9252--9265.

\bibitem{wacholder2014annotating}
{\sc Wacholder, N., Muresan, S., Ghosh, D., and Aakhus, M.}
\newblock Annotating multiparty discourse: Challenges for agreement metrics.
\newblock In {\em Proceedings of LAW VIII-The 8th Linguistic Annotation Workshop\/} (2014), pp.~120--128.

\bibitem{walters2018model}
{\sc Walters, E.}
\newblock The model rules of autonomous conduct: Ethical responsibilities of lawyers and artificial intelligence.
\newblock {\em Ga. St. UL Rev. 35\/} (2018), 1073.

\bibitem{wang2024benchmark}
{\sc Wang, A., Hertzmann, A., and Russakovsky, O.}
\newblock Benchmark suites instead of leaderboards for evaluating ai fairness.
\newblock {\em Patterns 5}, 11 (2024).

\bibitem{wei2024evaluating}
{\sc Wei, B., Shi, W., Huang, Y., Smith, N.~A., Zhang, C., Zettlemoyer, L., Li, K., and Henderson, P.}
\newblock Evaluating copyright takedown methods for language models.
\newblock Tech. rep., arXiv preprint arXiv:2406.18664, 2024.

\bibitem{weiser2023here}
{\sc Weiser, B.}
\newblock Here’s what happens when your lawyer uses chatgpt.
\newblock {\em The New York Times 27\/} (2023).

\bibitem{wen2024automate}
{\sc Wen-Yi, A.~W., Adamson, K., Greenfield, N., Goldberg, R., Babcock, S., Mimno, D., and Koenecke, A.}
\newblock Automate or assist? the role of computational models in identifying gendered discourse in us capital trial transcripts.
\newblock In {\em Proceedings of the AAAI/ACM Conference on AI, Ethics, and Society\/} (2024), vol.~7, pp.~1556--1566.

\bibitem{wendel2019promise}
{\sc Wendel, W.~B.}
\newblock The promise and limitations of artificial intelligence in the practice of law.
\newblock {\em Okla. L. Rev. 72\/} (2019), 21.

\bibitem{yamane2020artificial}
{\sc Yamane, N.}
\newblock Artificial intelligence in the legal field and the indispensable human element legal ethics demands.
\newblock {\em Geo. J. Legal Ethics 33\/} (2020), 877.

\bibitem{zamfirescu2022trucks}
{\sc Zamfirescu-Pereira, J., Chen, J., Wen, E., Koenecke, A., Garg, N., and Pierson, E.}
\newblock Trucks don’t mean {Trump}: Diagnosing human error in image analysis.
\newblock In {\em Proceedings of the 2022 ACM Conference on Fairness, Accountability, and Transparency\/} (2022), pp.~799--813.

\bibitem{zhao2025quantification}
{\sc Zhao, R., Choi, A.~S., Koenecke, A., and Rameau, A.}
\newblock Quantification of automatic speech recognition system performance on d/deaf and hard of hearing speech.
\newblock {\em The Laryngoscope 135}, 1 (2025), 191--197.

\end{thebibliography}

\end{document}